\documentclass[review]{elsarticle}
\usepackage{hyperref}

\usepackage{graphicx}
\usepackage{amsmath,amssymb,amsfonts,bm,mathtools,amsthm}
\usepackage{textcomp}
\usepackage[dvipsnames]{xcolor}
\usepackage{algorithm,algpseudocode}

\usepackage{graphbox}
\usepackage{subfig}
\usepackage{enumitem}
\usepackage{multirow}
\usepackage{makecell}
\usepackage{wrapfig}

\usepackage{lmodern}
\DeclareMathAlphabet{\fol}{OT1}{lmtt}{b}{n}

\newcommand{\thetab}{\bm\theta}

\newcommand{\x}{\mathbf{x}}
\newcommand{\bfb}{\mathbf{b}}
\newcommand{\bfc}{\mathbf{c}}
\newcommand{\bfd}{\mathbf{d}}

\newcommand{\bfh}{\mathbf{h}}
\newcommand{\bfu}{\mathbf{u}}
\newcommand{\bfv}{\mathbf{v}}
\newcommand{\bfw}{\mathbf{w}}
\newcommand{\bfx}{\mathbf{x}}
\newcommand{\bfy}{\mathbf{y}}

\newcommand{\I}{\mathbb{I}}

\newtheorem*{example*}{Example}{\bfseries}{\itshape}

\journal{Journal of Approximate Reasoning}

\begin{document}

\begin{frontmatter}

\title{Non-Parametric Learning of \\ Lifted Restricted Boltzmann Machines}


\author[mymainaddress]{Navdeep Kaur\corref{mycorrespondingauthor}}
\cortext[mycorrespondingauthor]{Corresponding author}
\ead{Navdeep.Kaur@utdallas.edu}
\author[mymainaddress]{Gautam Kunapuli}
\author[mymainaddress]{Sriraam Natarajan}

\address[mymainaddress]{The University of Texas at Dallas}



\begin{abstract}
We consider the problem of discriminatively learning restricted Boltzmann machines in the presence of relational data. Unlike previous approaches that employ a rule learner (for structure learning) and a weight learner (for parameter learning) sequentially, we develop a gradient-boosted approach that performs both simultaneously. Our approach learns a set of weak relational regression trees, whose paths from root to leaf are conjunctive clauses and represent the structure, and whose leaf values represent the parameters. When the learned relational regression trees are transformed into a lifted RBM, its hidden nodes are precisely the conjunctive clauses derived from the relational regression trees. This leads to a more interpretable and explainable model. Our empirical evaluations clearly demonstrate this aspect, while displaying no loss in effectiveness of the learned models. 
\end{abstract}


\end{frontmatter}

\section{Introduction}

Restricted Boltzmann Machines (RBMs, ~\cite{RBMFirstPaper1987}) have emerged as one of the most popular probabilistic learning methods. Coupled with advances in theory of learning RBMs: contrastive divergence (CD, ~\cite{ContrastiveDivergence2002}), persistent CD~\cite{PersistentCD2008}, and parallel tempering~\cite{paralleltempering2010} to name a few, their applicability has been extended to a variety of tasks~\cite{taylor06}. While successful, most of these models have been typically used with a flat feature representation (vectors, matrices, tensors) and not necessarily in the context of relational data. In problems where data is relational, these approaches typically flatten the data by either propositionalizing them or constructing embeddings that allowed them to employ standard RBMs. This results in the loss of ``natural'' interpretability that is inherent to relational representations, as well as a possible decline in performance due to imperfect propositionalization/embedding.

Consequently, there has been recent interest in developing neural models that directly operate on relational data. Specifically, significant research has been conducted on developing graph convolutional neural networks~\cite{GraphConvolutionalNetwork2018} that model graph data (a restricted form of relational data). Most traditional truly relational/logical learning methods~\cite{staraiBook,srlBook} are capable of learning with data of significantly greater complexity, including hypergraphs. Such representations have also been recently adapted to learning neural models~\cite{CollectiveClassificationPham2017,RelNNKazemi2018,LRNN2015}. One recent approach in this direction is Lifted RBMs~\cite{KaurEtAl18-RRBM}, where relational random walks were learned over data (effectively, randomized compound relational features) and then employed as input layer to an RBM.

While reasonably successful, this method still {\em propositionalized} relational features by constructing two forms of data aggregates: counts and existentials.  Motivated by this limitation, we propose a full, lifted RBM (LRBM), where the inherent representation is relational. Additionally, the LRBM can be learned without significant feature engineering, that is, a key component of our approach is discovering the structure of lifted RBMs. We propose a gradient-boosting approach for {\bf learning both the structure and parameters of LRBMs simultaneously}.
The resulting hidden nodes are newly discovered features, represented as conjunctions of logical predicates. 

These hidden layers are learned using the machinery of functional-gradient boosting~\cite{Friedman2001} on relational data. The idea is to learn a sequence of relational regression trees (RRTs) and then transform them to an LRBM by identifying appropriate transformations. There are a few salient features of our approach: (1) in addition to being well-studied and widely used~\cite{ImitationLearning2011, icdm11, rdnmlj11, TildeCRF2006}, RRTs can be parallelized and adapted easily to new, real-world domains; (2) our approach can handle hybrid data easily, which is an issue for many logical learners; (3) perhaps most important, our approach is {\bf explainable}, unlike other neural models. This is due to the fact that the hidden layers of the LRBM are simple conjunctions (paths in a tree), and can be easily interpreted as opposed to complex embeddings\footnote{Embedding approaches transform data from the input space to a feature space. A familiar example of this is {\em Principal Components Analysis}, which transforms input features to compound features via linear combination; the new features are no longer {\em naturally interpretable}. This is also the case with deep learning, which diminish interpretability by chaining increasingly complex feature combinations across successive layers (for example, autoencoders).}. Finally, (4) due to the nature of our learning method, we learn sparser LRBMs compared to employing random walks. 

We make a few key contributions in this work: (1) as far as we are aware, this is the first principled approach to learning truly lifted RBMs from relational data; (2) our representation ensures that the resulting RBM is interpretable and explainable (due to the hidden layer being simple conjunctions of logical predicates). We present (3) a gradient-boosting algorithm for simultaneously learning the structure and parameters of LRBMs as well as (4) a transformation process to construct a sparse LRBM from an ensemble of relational regression trees produced by gradient boosting. Finally, (5) our empirical evaluation clearly demonstrates three aspects: efficacy, efficiency and explainability of our approach compared to the state-of-the-art on several data sets.

\section{Background and Related Work}
Scalars are denoted in lower-case ($y$, $w$), vectors in bold face ($\bfy$, $\bfw$), and matrices in upper case ($Y$, $W$). $\bfu ^\intercal \bfv$ denotes the dot product between $\bfu$ and $\bfv$.

\paragraph{Restricted Boltzmann Machines} RBMs are stochastic neural networks 
%
%
%
\begin{figure}[!thb]
    \centering
    \includegraphics[width=0.675\textwidth]{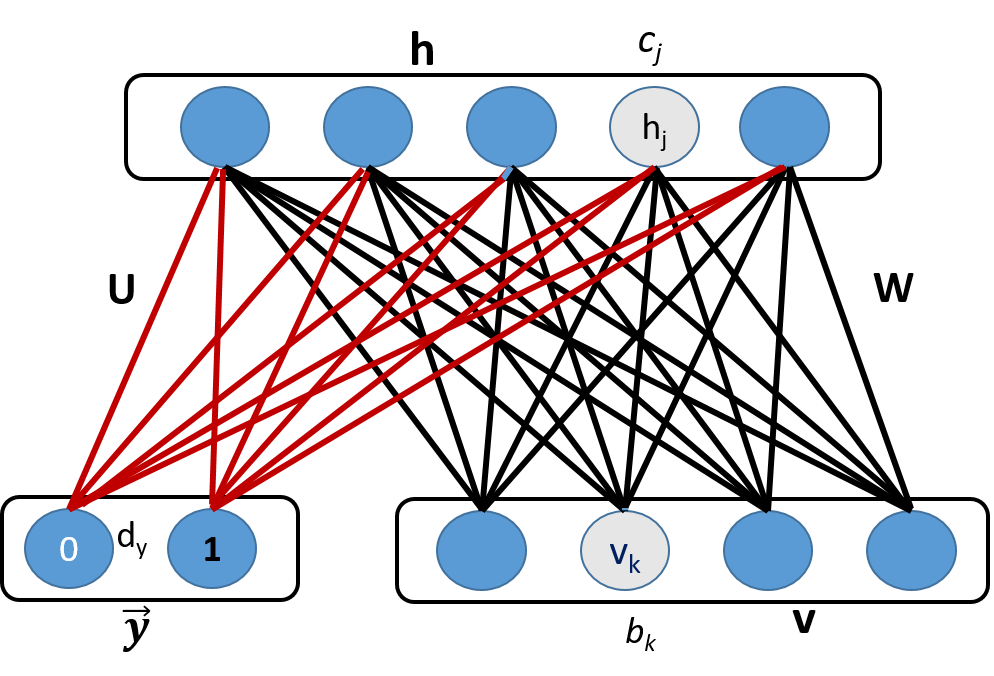}
    \caption{A discriminative RBMs has a dense set of connections between the visible and hidden layers. This figure illustrates a discriminative RBM for a binary classification problem, though this model can naturally handle multi-class problems by adding additional output nodes (corresponding to the one-hot vectorization of the label).}
    \label{fig: RBMFigure}
\end{figure}
consisting of a {\em hidden layer} of neurons that model the probability distribution of a {\em visible layer} of neurons. Specifically, discriminative RBMs~\cite{DiscrRBM2008} (Figure \ref{fig: RBMFigure}) have a Bernoulli input layer (also known as the visible layer, $\bfv$), a Bernoulli hidden layer ($\bfh$) and a softmax output layer ($\bfy$). The joint configuration ($\bfy, \bfv, \bfh$) of the model has the following energy:

\begin{equation} 
 E(\bfy,\bfv, \bfh) = -\bfh^\intercal W\mathbf{v} - \mathbf{b}^\intercal \mathbf{v} - \mathbf{c}^\intercal\bfh - \bfd^\intercal \bfy - \bfh^\intercal U\bfy,
 \label{eq: energyequation}
\end{equation}

where $W$ are the weights between the visible and hidden layer, $ U$ are the weights between the hidden and the output layer, and $\bfb$, $\bfc$, $\bfd$ are the biases in the visible, hidden and output layers respectively. Given a (multi-class) label $y=\ell$, $\ell \in \{1, \, \hdots, \, C \}$, the output is a one-hot vector $\bfy = (I_{c=\ell})_{c=1}^{C}$. With a slight abuse of notation, we denote the multi-class label of a training example as $y$, with its corresponding vectorization in bold as $\bfy$. The joint probability distribution of the RBM can be written as: $P(\bfy,\bfv, \bfh) = \frac{1}{Z} e^{-E(\bfy,\bfv, \bfh)}$,
where $Z$ is normalization constant. While computing $P(\bfy, \bfv, \bfh)$ is generally intractable, the conditional, $P(y \mid \bfv)$, can be computed exactly:
\begin{equation}
    p(y \mid\bfv) = \frac{\exp \left( d_{y} + \sum_{j} \, \zeta(c_{j} + U_{jy} + \sum_{k} W_{jk} v_{k}) \right) }{\displaystyle{\sum_{y* \in \{1,2,..C \}}} \,  \exp \left( d_{y*} + \textstyle{\sum_{j}} \, \zeta(c_{j} + U_{jy*} + \sum_{k} W_{jk} v_{k}) \right)},
    \label{eq: discRBM}
\end{equation}
where $\zeta(a) = \log(1 + e^a)$, the {\em softplus function}. Our goal is to extend this formulation to relational domains and learn the resulting {\bf Lifted Restricted Boltzmann Machines} (LRBMs) using functional gradient boosting. 

\paragraph{Functional Gradient Boosting} 
Functional gradient boosting (FGB), introduced by Friedman \cite{Friedman2001} in 2001, has recently emerged as a state-of-the-art ensemble method. Functional gradient boosting aims to learn a model $f(\cdot)$ by optimizing a loss function $\mathcal{L}[f]$ by emulating gradient descent. At iteration $m$, however, instead of explicitly computing the gradient $\partial L[f_{m-1}](\x_i, y_i)$, FGB approximates the gradient using a weak regression tree\footnote{A weak base estimator is any model that is ``simple'' and underfits (hence, weak). From a machine-learning standpoint, such weak learners are high bias, low variance and easy to learn. Shallow decision trees are an exceptionally popular choice for weak base estimators for ensemble learning, owing to their algorithmic efficiency and interpretability.}, $\Delta_m$.

For a probabilistic model, the loss function is replaced by a (log-)likelihood function ($L[\psi]$), which is described in terms of a potential function $\psi(\cdot)$, which FGB aims to learn. FGB begins with an initial potential $\psi_0$; intuitively, $\psi_0$ represents the prior of the probability distribution of target atom. This initial potential can be any function: a constant, a prior probability distribution or any function that incorporates background knowledge available prior to learning. 

\begin{figure}[!t]
\begin{center}
    \includegraphics[scale=0.6]{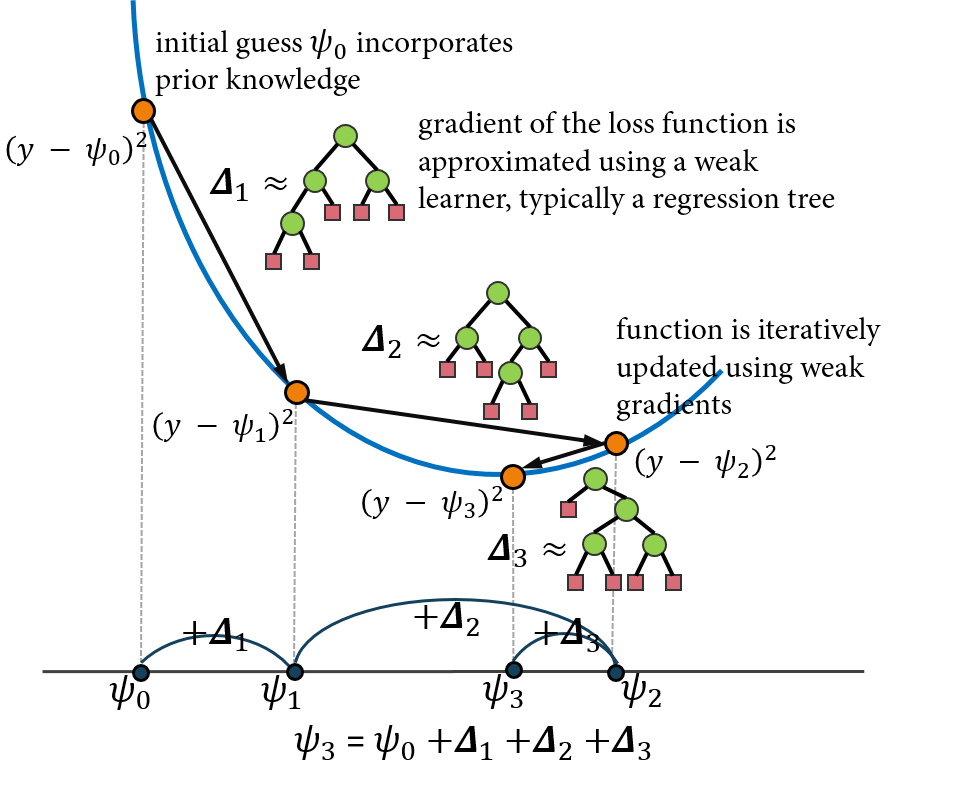}
\end{center}
\vspace{-0.25in}
\caption{Functional Gradient Boosting, where the loss function is mean squared error.}
\label{fig: FGB Explained}
\end{figure}

At iteration $m$, FGB approximates the true gradient by a {\em functional gradient} $\Delta_m$. That is, gradient boosting will attempt to identify an approximate gradient $\Delta_m$ that corrects the errors of the current potential, $\psi_{m-1}$. This ensures that the new potential $\psi_m = \psi_{m-1} + \Delta_m$ continues to improve. Like most boosting algorithms, FGB learns $\Delta_m$ as a weak regression tree, and ensembles several such weak trees to learn a final potential function (see Figure \ref{fig: FGB Explained}). Thus, the final model is a sum of regression trees $\psi_{m} = \psi_{0} + \Delta_{1} + \hdots + \Delta_{m}$ (Figure \ref{fig: FGB Explained}). 

In relational models, regression trees are replaced by {\em relational regression trees} (RRTs, \cite{TildeBlockeel1998}). This allows us to learn relational conditional models such as Relational Dependency Networks \cite{rdnmlj11}, Relational Logistic Regression~\cite{RLRBoost2018}, relational policies~\cite{ImitationLearning2011}, discriminative training of undirected models~\cite{icdm11} and even temporal models~\cite{Yang2016AAAI}. Inspired by these methods, we propose to learn the hidden layer of an LRBM using gradient boosting.


\paragraph{Relational Neural Models} Relational Embeddings~\cite{RESCAL2011,TransE2013,NTN2013,DistMult2015,HolE2016,complex2016} have gained popularity recently. A common theme among current approaches is to learn a vector representation, that is, an embedding for each relation and each entity present in the knowledge base. Most of these approaches also assume binary relations, which is a rather restrictive assumption that cannot capture the richness of  real-world relational domains. Further, they need a large number of embeddings for training, especially the deep-learning-based approaches. Finally, and possibly most concerning: many embedding approaches cannot easily generalize to new data, and the entire set of embeddings has to be relearned  with new data, or for every new task.
%


Approaches closest to our proposed work are relational neural networks \cite{RelNNKazemi2018,LRNN2015,CLIPPLusPlus2014,DiMaio2004,DRM2013}; these approaches also represent the structure of a neural network as first-order clauses as we do. The key difference however, is that in all these models, clauses have already been obtained either from an expert or an independent ILP system. That is to say, domain rules that make up its structure and the resulting neural network architectures are manually specified, and these approaches typically only perform parameter learning. 

Recently, relational neural networks have been proposed for vision tasks \cite{DeepMindNN2017,FewShotLearning2018,RelNNforObjectDetection2018}. While promising, these networks have fixed, manually-specified structures and the nature of the relations captured between objects is also not interpretable or explainable.
In contrast, our model learns the structure and parameters of neural network {\bf simultaneously}. One common theme among all these models is that they learn latent features of relational data in their hidden layers, but our model, being still in its nascent stage, cannot do so yet.

A few approaches for learning neural network on graphs exist. Graph convolutional networks \cite{NiepertGraphConNet2016} enable graph data to be trained directly on convolutional networks. 
Another set of popular approaches \cite{GraphNeuralNetwork2009} train a recurrent neural network on each node of the graph by accepting the input from neighboring nodes until a fixed point is reached. The work of Scarcelli et al. \cite{GraphNeuralNetwork2009} extends this by learning embeddings for entities and relations in the relational graph. 

Recently, Pham et al. \cite{CollectiveClassificationPham2017} proposed a neural network architecture where connections in the different nodes of network are encoded according to given graph structure. RBMs have also been considered in the context of relational data. For instance, two tensor based models \cite{GatedCHBM2015,LRBM2014} proposed to lift RBMs by incorporating a four-order tensor into their architecture that captures interaction between quartet consisting of two objects, relation existing between them and hidden layer. Finally, our recent approach \cite{KaurEtAl18-RRBM} learns relational random walks and uses the counts of the groundings as observed layer of an RBM. 



\section{Boosting of Lifted RBMs}

Recall that our goal is to learn a {\em truly lifted RBM}. Consequently, both the hidden and observed layers of the RBM should be lifted (parameterized as against propositional RBMs). This is to say that, the observed layers are the predicates (logical relations describing interactions) in the domain, while the hidden layer consists of conjunctions of predicates (logical rules) learned from data. Instead of a complete network, connections exist only between predicates and hidden nodes that are present in the conjunction. We illustrate RBM lifting with the following example.
\begin{example*}
Consider a movie domain that contains the entity types (variables) $\fol{Person}(\fol{P})$, $\fol{Movie}(\fol{M})$ and $\fol{Genre}(\fol{G})$. Predicates in this domain describe relationships between the various entities, such as $\fol{DirectedBy(\fol{M},\fol{P})}$, $\fol{ActedIn(\fol{P},\fol{M})}$, $\fol{InGenre(M, G)}$ and entity resolution predicates such as $\fol{SamePerson(P_1,P_2)}$ and $\fol{SameGenre(G_1,G_2)}$. These predicates are the {\em atomic domain features}, $f_i$. The task is to predict the nature of the collaboration between two persons $\fol{P_1}$ and $\fol{P_2}$; this task can be represented via the target predicate:
\[
\fol{Collaborated(P_1,P_2)} \, = \, \left\{
\begin{array}{cl}
    0, & \fol{P_1}, \fol{P_2} \,\, \textrm{never collaborated}, \\
    1, & \fol{P_1} \,\, \textrm{worked under} \,\, \fol{P_2},  \\
    2, & \fol{P_2} \,\, \textrm{worked under} \,\, \fol{P_1}, \\
    3, & \fol{P_1}, \fol{P_2} \,\, \textrm{collaborated at the same level}.
\end{array}
\right. 
\]
To perform this $4$-class classification task, we can construct more complex lifted features through conjunctions of the atomic domain features. For example, consider the following lifted feature, $h_1$:
\begin{equation}
\left( \begin{array}{r}
    \fol{DirectedBy(M_1, P_1) \wedge InGenre(M_1, G_1) \, \wedge } \\
    \fol{ActedIn(P_2, M_2) \wedge InGenre(M_2, G_2) \, \wedge }\\ 
    \neg \, \fol{SameGenre(G_1, G_2)} \,\,\,\,\,\,\\
\end{array}\right) \, \Rightarrow \, \fol{\left( \,  Collab(P_1, P_2) = 0 \, \right)}.
\tag{$h_1$}
\end{equation}
This lifted feature is a {\em compound domain rule} (essentially a typical conjunction in logic models) made up of several atomic domain features that describes one possible classification condition of the target predicate. Specifically, the lifted feature $h_1$ expresses the situation where two persons $\fol{P_1}$ and $\fol{P_2}$ are unlikely to have collaborated if they work in different genres. Every such compound domain rule becomes lifted feature with a corresponding hidden node. In this example, we introduce two others:
\[
\begin{array}{lc}
\fol{DirBy(M_1, P_1) \wedge ActedIn(P_3, M_1) \wedge SamePer(P_3, P_2)} \Rightarrow \fol{\left( \, Collab(P_1, P_2) = 1 \, \right), } &
(h_2)\\
\fol{ActedIn(P_1, M) \wedge ActedIn(P_2, M)} \Rightarrow \fol{\left( \, Collab(P_1, P_2) = 3 \, \right)}. &
(h_3)
\end{array}
\]

\begin{figure}[t]
    \centering
    \includegraphics[scale=0.75]{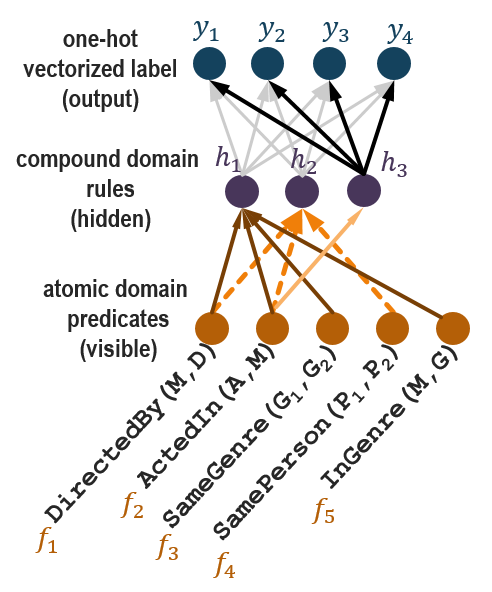}
    \caption{An example of a lifted RBM. The atomic predicates each have a corresponding node in the visible layer ($f_i$). Atomic predicates can be used to create richer features as conjunctions, which are represented as hidden nodes ($h_j$); the connections between the visible and hidden layers are sparse and only exist when the predicate corresponding to $f_i$ appears in the compound feature $h_j$. The output layer is a one-hot vectorization of a multi-class label $\bfy$, and has one node for each class $y_k$. The connections between the hidden and output layers are dense and allow all features to contribute to reasoning over all the classes.}
    \label{fig: example lifted RBM}
\end{figure}

The key intuition is that these rules, or lifted features, capture the latent structure of the domain and are a critical component of lifting RBMs. The layers of the lifted RBM are as follows (Figure \ref{fig: example lifted RBM}):
\begin{itemize}
\itemsep0em
    \item {\bf Visible layer, atomic domain predicates}: We create a visible node $v_i$ for each lifted atomic domain predicate $f_i$. Thus, we can express any possible structure that can be enumerated as conjunction of these atomic features. In Figure \ref{fig: example lifted RBM}, the visible layer consists of the five atomic predicates introduced above, $f_1, \, \hdots, \, f_5$.
    
    \item {\bf Hidden layer, compound domain rule}: Each of the compound features can be represented as a node in the hidden layer, $h_i$. In this manner, the lifted RBM is able to construct and use complex structural rules to reason over the domain. This is similar to classical neural networks, propositional RBMs and deep learning, where the hidden layer neurons represent rich and complex feature combinations. 
    
    The key difference from existing architectures is that the connections between the visible and hidden layers are {\bf not dense}; rather, they are extremely sparse and depend only on the atomic predicates that appear in the corresponding lifted compound features. In Figure \ref{fig: example lifted RBM}, the hidden node $h_1$ is connected to the atomic predicate nodes $f_1$, $f_2$, $f_3$ and $f_5$, while the hidden node $h_3$ is connected to only the atomic predicate node $f_2$. This allows the lifted RBM to represent the domain structure in a compact manner. Furthermore, such ``compression'' can enable acceleration of weight learning as unnecessary edges are not introduced into the structure.
    
    \item {\bf Output layer, one-hot vectorization}: As mentioned above, the lifted RBM formulation can easily handle multi-class classification. In this example, the target predicate can take $4$ values as it corresponds to a $4$-class classification problem. This can be modelled with four output nodes $y_1, \hdots, \, y_4$ through one-hot vectorization of the labels. Note that the connections between the hidden and output layers {\bf are dense}. This is to ensure that all features can contribute to the classification of all the labels.
    
    Furthermore, this enables the lifted RBM to reason with uncertainty. For example, consider the compound domain feature $h_1$, which describes a condition for two persons to have never collaborated. By ensuring that the hidden-to-output connections are dense, we allow for the contribution of this rule to the final prediction to be {\em soft} rather than {\em hard}. This is similar to how Markov logic networks learn different rule weights to quantify the relative importance of the domain rules/lifted features. In a similar manner, the lifted RBM allows for reasoning under uncertainty by learning the network weights to reflect the relative significance of various features to different labels.
\end{itemize}
\end{example*}


Our task now is to learn such lifted RBMs. Specifically, we propose to learn the structure (compound features as hidden nodes) as well as the parameters (weights on all the edges and biases within the nodes). This is a key novelty as our approach uses gradient boosting to learn sparser LRBMs, unlike the fully connected propositional ones. To learn an LRBM, we need to (1) formulate the (lifted) potential definitions, (2) derive the functional gradients, (3) transform the gradients to {\bf explainable} hidden units of the RBM, and (4) learn the parameters of the RBM. We now present each of these steps in detail. 

\subsection{Functional Gradient Boosting of Lifted RBMs}
The conditional equation (\ref{eq: discRBM}) which is the basis of an RBM, is formulated for propositional data, where each feature of a training example $\bfx_i$ is modeled as a node in the input layer $\bfv$. We now extend this definition of the RBM to handle logical predicates (i.e., parameterized relations). 


Note that these lifted features (conjunctions) can be obtained in several different ways: (i) as with many existing work on neuro-symbolic reasoning, these could be provided by a domain expert, or (ii) can be learned from data similar to the research inside Inductive Logic Programming \cite{Muggleton94} or, (iii) performing random walks in the domain that result in rule structures~\cite{KaurEtAl18-RRBM}, to name a few. Any rule induction technique could be employed in this context. In this work, we adapt a gradient-boosting technique. Given such lifted features (or rules) $f_k(\bfx)$ on training examples $\bfx$, we can rewrite equation (\ref{eq: discRBM}) as 
\begin{equation}
    p(y \mid \x) = \frac{\exp \left( d_{y} + \sum_{j} \, \zeta(c_{j} + U_{jy} + \sum_{k} W_{jk} f_{k}(\x)) \right) }{\displaystyle{\sum_{y* \in \{1,2,..C \}}} \,  \exp \left( d_{y*} + \textstyle{\sum_{j}} \, \zeta(c_{j} + U_{jy*} + \sum_{k} W_{jk} f_{k} (\x) ) \right)}.
    \label{eq: RelationaldiscRBM}
\end{equation}
Contrast this expression to the propositional discriminative RBM (equation \ref{eq: discRBM}), which models $p(y \mid \bfv)$. The key difference is that the propositional features $\sum_{k} W_{jk} v_{k}$ are replaced with lifted features $\sum_{k} W_{jk} f_{k}(\x)$; while features in a propositional data set are just the data columns/attributes, the features in a relational data set are typically represented in predicate logic (as shown in the example above) and are rich and expressive conjunctions of objects, their attributes and relations between them.

We now introduce some additional functional notation to simplify (equation \ref{eq: RelationaldiscRBM}). %
Without loss of generality, we restrict our discussion to the case of binary targets (with labels $\ell \in \{ 0, \, 1\}$) and note that this exposition can easily be extended to the case of multiple classes. For each label $\ell$, we define functional
\[
E(\x_i \mid \bfc, W, \bfd_\ell, U_\ell) \, \coloneqq \, d_\ell \, + \, \sum_{j} \, \zeta \left( c_{j} + U_{j\ell} + \sum_{k} W_{jk} f_{k}(\x_i) \right).
\]
This functional represents the ``energy'' of the combination $(\x_i, \, y_i = \ell)$. For binary classification, (equation $\ref{eq: RelationaldiscRBM}$) is further simplified to
\begin{equation}
p(y_i=1 \mid \x_i) \, = \, \frac{e^{E(\x_i \mid \bfc, W, d_1, U_1)}}{e^{E(\x_i \mid \bfc, W, d_0, U_0)} + e^{E(\x_i \mid \bfc, W, d_1, U_1)}}. 
\label{eq: DiscriminativeRBM-short}
\end{equation}
This reformulation is critical for the extension of the discriminative RBM framework to relational domains as it allows us to rewrite the probability $p(y_i=1 \mid \x_i)$ in terms of a functional that represents the potential and $\textsf{OF}(\x_i)$, the observed features of the training example $\x_i$. One of our goals is to learn lifted features from the set of all possible features. In simpler terms, if $\x$ is the set of all predicates in the domain and $x$ is the current target, then the goal is to identify the set of features $\textsf{OF}(x)$ s.t, $P(x \mid \x) = P(x \mid \textsf{OF}(x))$. In Markov network terminology, this refers to the Markov blanket of the corresponding variable. In a discriminative MLN framework, $\textsf{OF}(x)$ is the set of weighted clauses in which the predicate $x$ appears. We can now define the probability in (equation \ref{eq: DiscriminativeRBM-short}) as 
\begin{eqnarray}
    p_\psi \left( y_i = 1 \mid \textsf{OF}(\x_i) \right) = \frac{e^{\psi(y_i=1 \mid \textsf{OF}(\x_i))}}{1 + e^{\psi(y_i=1 \mid \textsf{OF}(\x_i))}}, \,\,\,\ \textrm{where}\\
    \label{eq: ProbEqn}
    \psi(y_i = 1 \mid \textsf{OF}(\x_i)) \, = \, E(\x_i \mid \bfc, W, d_1, U_1) - E(\x_i \mid \bfc, W, d_0, U_0).
\label{eq: potential as difference}
\end{eqnarray}
Note that $\textsf{OF}(\bfx_i)$ does not include {\em all} the features in the domain, but only the specific features that are present in the hidden layer. An example of this can be observed in Figure~\ref{fig: example lifted RBM}. This LRBM consists of three lifted features $\langle h_1,h_2,h_3 \rangle$ that correspond to the three rules mentioned earlier. We can thus explicitly write the potential function for a lifted RBM (equation \ref{eq: potential as difference}) in functional form as\begin{equation}
\psi(y_i = 1 \mid \textsf{OF}(\x_i)) = d + \sum_{j} \, \log \left( \frac{1 +  \exp{\left(c_{j} + U_{j1} + \sum_{k} W_{jk} f_{k}(\x_i)\right)}}{1 + \exp{ \left( c_{j} + U_{j0} + \sum_{k} W_{jk} f_{k}(\x_i) \right)}} \right),
\label{eq: PsiDefinition}
\end{equation}
\begin{wrapfigure}{r}{0.5\textwidth}
\vspace{-0.2in}
\begin{center}
    \includegraphics[scale=0.8]{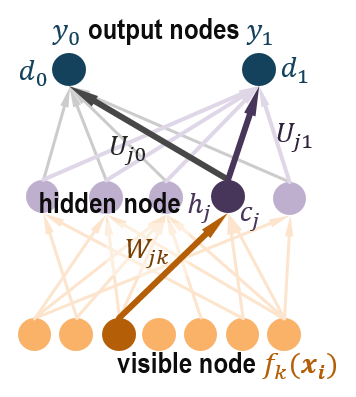}
\end{center}
\vspace{-0.25in}
\caption{Weights in a lifted RBM.}
\label{fig: LRBM weights}
\end{wrapfigure}
where $d = d_1 - d_0$. This potential functional is parameterized by $\thetab \, = \, \{ d, \bfc, W, U_0, U_1 \}$, consisting of (see Figure \ref{fig: LRBM weights}) edge weights and biases. The edge weights to be learned are $W_{jk}$ (between visible node corresponding to feature $f_k(\bfx_i)$ and hidden node $h_j$) and $U_{j\ell}$ (between hidden node $h_j$ and output node $y_\ell$). The biases to be learned are $c_j$ on the hidden nodes and $d_\ell$ on the output nodes. However, instead of learning two biases $d_1$ and $d_0$, we can learn a single bias $d = d_1 - d_0$ as the functional $\psi$ only depends on the difference (see equation \ref{eq: PsiDefinition}). Given this functional form, we can now derive a functional gradient that maximizes the overall log-likelihood of the data
\[
L(\{\x_i, \, y_i\}_{i=1}^n \mid \psi) \, = \, \log \prod_{i=1}^n p_\psi \left( y_i = 1 \mid \textsf{OF}(\x_i) \right) \, = \, \sum_{i=1}^n
\, \log p_\psi \left( y_i = 1 \mid \textsf{OF}(\x_i) \right). \]
The (pointwise) functional gradient of $L(\{\x_i, \, y_i\}_{i=1}^n \mid \psi)$ with respect to $\psi(y_i = 1 \mid \textsf{OF}(\x_i))$ can be computed as follows,
\[
\frac{\partial \log p_\psi(y_i = 1 \mid \textsf{OF}(\x_i))}{\partial \psi(y_i = 1 \mid \textsf{OF}(\x_i))} \, = \, \I(y_i = 1) - P(y_i = 1 \mid \textsf{OF}(\x_i)) \, \coloneqq \,  \Delta_i, 
\]
where $\I(y_i = 1)$ is an indicator function. The pointwise functional gradient has an elegant interpretation. For a positive example ($\I(y_i = 1) = 1$), the functional gradient $\Delta_i$ aims to improve the model such that $1 - P(y_i = 1)$ is as small as possible, in effect pushing $P(y_i = 1) \rightarrow 1$. For a negative example, ($\I(y_i = 1) = 0$), the functional gradient $\Delta_i$ aims to improve the model such that $0 - P(y_i = 1)$ is as small as possible, in effect pushing $P(y_i = 1) \rightarrow 0$. Thus, the gradient of each training example $\x_i$ is simply the adjustment required for the probabilities to match the true observed labels $y_i$. The functional gradient derived here has a similar form to the functional gradients in other relational tasks such as boosting relational dependency networks \cite{rdnmlj11} Markov logic networks \cite{icdm11} and relational policies~\cite{ImitationLearning2011} to specify a few.

\subsection{Representation of Functional Gradients for LRBMs}
Our goal now is to approximate the true functional gradient by fitting a regression function $\hat \psi(\x)$ that minimizes the squared error over the pointwise gradients of all the individual training examples:
\begin{equation}
     \hat \psi (\x) \, = \, \arg\min_{\psi} \, \sum_{i=1}^{n} (\psi(\x_i \mid \textsf{OF}(\x_i)) - \Delta_i)^2.
     \label{eq: SSEEquation}
\end{equation}

We consider learning the representation of $\hat{\psi}$ as a sum of relational regression trees. The key advantage is that a relational tree can be easily interpreted and explained. To learn a tree to model the functional gradients, we need to change the typical tree learner. Specifically, the splitting criteria of the tree at each node is modified; to identify the next literal to add to the tree, $\fol{r(x)}$, we greedily search for the literal that minimizes the squared error (equation \ref{eq: SSEEquation}).


\begin{figure}
    \centering
    \includegraphics[scale=0.75]{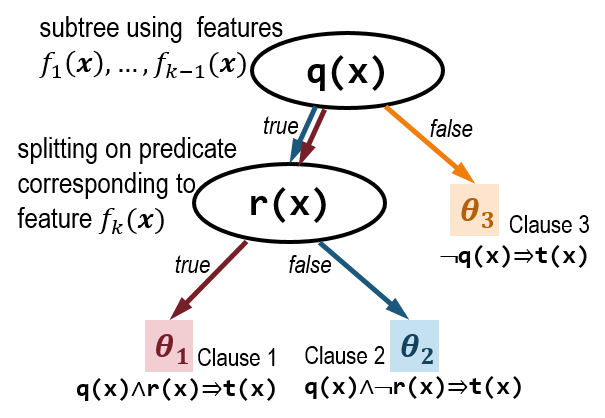}
    \caption{A general relational regression tree for lifted RBMs when learning a target predicate $\fol{t(x)}$. Each path from root to leaf is a compound feature (also a logical clause $\mathtt{Clause}_r$) that enters the RBM as a hidden node $h_r$. The leaf node contains the weights $\thetab_r \, = \, \{ d^r, \bfc^r, W^r, U^r_0, U^r_1 \}$ of all edges introduced into the lifted RBM when this hidden node/discovered feature is introduced into the RBM structure.}
    \label{fig: LRBM RRT}
\end{figure}

For a tree-based representation, we employ a relational regression-tree (RRT) learner \cite{TildeBlockeel1998} to learn a function to approximately fit the gradients on each example. If we learn an RRT to fit $\psi(\x_i \mid \mathsf{OF}(\x_i))$ in equation (\ref{eq: PsiDefinition}), each path from the root to a leaf can be viewed as a clause, and the leaf nodes correspond to an RBM that evaluates to the weight of the clause for that training example. Figure \ref{fig: LRBM RRT} shows an RRT when learning a lifted RBM via gradient boosting for some target predicate $\fol{t(x)}$. The node $\fol{q(x)}$ can be any subtree that has been learned thusfar, and a new predicate $\fol{r(x)}$ has been identified as a viable candidate for splitting. On splitting, we obtain two new compound features as evidenced by two distinct paths from root to the two new leaf nodes. These clauses (paths) along with their corresponding leaf nodes identify a new structural component of the RBM, along with corresponding parameters
\[
\begin{array}{cl}
    (\fol{Clause_1})& \thetab_1: \fol{q(x) \wedge r(x) \Rightarrow t(x)}, \\
    (\fol{Clause_2})& \thetab_2: \fol{q(x) \wedge \neg r(x) \Rightarrow t(x)}.
\end{array}
\]
Note that the clause $\fol{q(\cdot)}$ and the predicate $\fol{r(\cdot)}$ are expressed generally, and their arguments are denoted broadly as $\fol{x}$. In practice, $\fol{q(\cdot)}$ and $\fol{r(\cdot)}$ can be of different arities and take any possible entity types in the domain.

\subsection{Learning Relational Regression Trees}
Let us assume that we have learned a relational regression tree till $\fol{q(x)}$ in Figure \ref{fig: LRBM RRT}. Now assume that, we are adding a literal $\fol{r(x)}$ to the tree at the left-most node of the subtree $\fol{q(x)}$.

Let the feature corresponding to the left branch ($\fol{Clause_1}$) be $f_1(\bfx) = \I(\fol{q(x) \wedge r(x)})$, that is, feature $f_1(\bfx) = 1$ for all training examples $\bfx$ that end up at the leaf $\thetab_1$ and zero otherwise. Similarly, let the feature corresponding to the right branch ($\fol{Clause_2}$) be $f_2(\bfx) = \I(\fol{q(x) \wedge \neg r(x)})$. The potential function $\psi(y_i = 1 \mid \textsf{OF}(\x_i))$ can be written using (equation \ref{eq: PsiDefinition}) as:
\begin{equation}
   \psi(y_i = 1 \mid \textsf{OF}(\x_i)) \, = \prod_{k=1, 2} \left[ d^k + \log \left( \frac{1 +  \exp{\left(c^k + U_1^k + W^k f_k(\x_i) \right)}}{1 + \exp{ \left( c^k + U_{0}^k + W^k f_k(\x_i) \right)}} \right)  \right]^{f_k(\x_i)} \hspace{-0.3in}.
   \label{eq: SSEeqwithpredicatelevelparameters}
\end{equation}
In this expression, when a training example $\bfx_i$ satisfies $\fol{Clause_1}$, it reaches leaf node $\thetab_1$ and consequently, we have $f_1(\bfx_i) = 1$ and $f_2(\bfx_i) = 0$. When a training example $\bfx_i$ satisfies $\fol{Clause_2}$, the converse is true and we have $f_1(\bfx_i) = 0$ and $f_2(\bfx_i) = 1$. Thus, only one term is active in the expression above and delivers the potential corresponding to whether the training example $\bfx_i$ is classified to the left leaf $\thetab_1$ or the right leaf $\thetab_2$. We can now substitute this expression for the potential in equation (\ref{eq: SSEeqwithpredicatelevelparameters}) into the loss function (\ref{eq: SSEEquation}).



The loss function is used in two ways to grow the RRTs:
\begin{enumerate}
    \item First, we identify the next literal to add to the tree, $\fol{r(x)}$, by greedily searching for the atomic domain predicate that minimizes the squared error. This is similar to the splitting criterion used in other lifted gradient boosting models such as MLN-boosting \cite{icdm11}. 
    \item Next, after splitting, we learn parameters for the newly introduced leaf nodes. That is, for each split of the tree at $\fol{r(x)}$, we learn $\thetab_1 = [d^1, c^1, W^1,$ $U^1_0, U^1_1]$ for the left subtree and $\thetab_2 = [d^2, c^2, W^2, U^2_0, U^2_1]$ for the right subtree. We perform parameter learning via coordinate descent \cite{Wright15-CoordinateDescent}.
\end{enumerate}

\subsection{LRBM-Boost Algorithm:}

\begin{algorithm}
\caption{{\tt LRBM-Boost}: Relational FGB for Lifted RBMs}
\begin{algorithmic}[1]
\Function{\sc LRBM-Boost}{$Data$, $T$, $N$}
\State F$_{0}$ = $\psi_{0}$ \Comment{\color{NavyBlue}  set prior of potential function}
\For {$1 \leq n \leq N$} 
    \State S := $\Call{GenerateExamples}{F_{n-1}, Data, T}$ \Comment{\color{NavyBlue}  examples for next tree}
    \State $F_{n}$ := $\Call{FitRegressionTree}{S, L, T}$
    \Comment{\color{NavyBlue}  learn regression tree}
    \State $F_{n}$ = $F_{n}$ + $F_{n-1}$ \Comment{\color{NavyBlue}  add new tree to existing set}
\EndFor
\State P(y$_{i}$ = 1 $\vert$ OF($\textbf{x}_{i}$)) $\propto$ exp($\psi(y_{i}$ = 1$\vert$ OF($\textbf{x}_{i}$)))
\EndFunction
\Function{\sc FitRegressionTree}{$S$, $L$, $T$}
\State $Tree$ := $\Call{CreateTree}{T(X)}$ \Comment{\color{NavyBlue}  create empty Tree}
\State $Beam$ := $\{Root(Tree)\}$
\While {($i \leq L$)} \Comment{\color{NavyBlue}  till max clauses $L$ is reached}
\State $N$ := $\Call{CurrentNodeToExpand}{Beam}$
\State $C$ := $\Call{GeneratePotentialChildren}{N}$
\For {{\bf each} $c$ in $C$} \Comment{\color{NavyBlue}  greedily search best child}
\State $\lbrack$ S$_{L}$, $\Delta_{L}$ $\rbrack$ := $\Call{ExampleSatisfaction}{N \wedge c, S}$    \Comment{\color{NavyBlue}  left subtree}
\State $\Theta_{L}^{c}$ := $\Call{CoordinateDescent}{\lbrack S_{L}, \Delta_{L}\rbrack}$ \Comment{\color{NavyBlue}  learn LRBM params}
\State $\lbrack$ S$_{R}$, $\Delta_{R}$ $\rbrack$ := $\Call{ExampleSatisfaction}{N \wedge \neg c, S}$ \Comment{\color{NavyBlue}  right subtree}
\State $\Theta_{R}^{c}$ := $\Call{CoordinateDescent}{\lbrack S_{R}, \Delta_{R}\rbrack}$ \Comment{\color{NavyBlue}  learn LRBM params}
\State $score_{c}$ := $\Call{ComputeSSE}{\Theta_{L}^{c}, \Theta_{R}^{c}, \Delta_{L}, \Delta_{R}}$ \Comment{\color{NavyBlue}  using eq.(\ref{eq: SSEEquation})}
\EndFor
\State $\hat{c}$ := $\underset{c}{argmin}$($score_{c}$) 
\State $\Call{AddChild}{Tree, N, \hat{c}}$ 
\State $\Call{Insert}{Beam, \hat{c}.left, \hat{c}.left.score}$
\State $\Call{Insert}{Beam, \hat{c}.right, \hat{c}.right.score}$
\EndWhile
\State \Return Tree
\EndFunction
\end{algorithmic}
\label{alg: RRBMBoostAlgo}
\end{algorithm}

We now describe \textsc{LRBM-Boost} algorithm (Algorithm \ref{alg: RRBMBoostAlgo}) to learn structure and parameters of LRBM. The algorithm takes instantiated ground facts ($Data$) and training examples of target $T$ as input and learns $N$ regression trees that fits the example gradients to set of trees. The algorithm starts by considering prior potential value $\psi_{0}$ in $F_{0}$, and in order to learn a new tree, it first generates the regression examples, $S$=$\lbrack$($\x_i$,$y_{i}$), $\Delta_{i}\rbrack$, in (line 4) where regression value $\Delta_{i}$ ($I$-$P$) is computed by performing inference of previously learned trees. These regression examples $S$ serve as input to \textsc{FitRegressionTree} function (line 5) along with the maximum number of leaves ($L$) in the tree. The next tree $F_{n}$ is then added to set of existing tree (line 6). The final probability of LRBM can be computed by performing inference on all $N$ trees in order to obtain $\psi$. 

$\textsc{FitRegressionTree}$ function (line 10) generates a relational regression tree with $L$ leaf nodes. It starts with an empty tree and greedily adds one node at a time to the tree. In order to add next node to the tree, it first considers the current node $N$ to expand as the one that has the best score in the beam (line 14). The potential children $C$ of this node $N$ (line 15) are constructed by greedily considering and scoring clauses where the parameters are learned using coordinate descent. Once the $\hat{c}$ is determined, it is added as the leaf to the tree and the process is repeated.


\section{Experimental Section}
\noindent We aim to answer the following questions in our experiments: 
\begin{itemize}
    \itemsep0em
    \item [{\bf Q1}] How does $\mathtt{LRBM}$-$\mathtt{Boost}$\footnote{ \url{https://github.com/navdeepkjohal/LRBM-Boost}} compare to other relational neural models? 
    \item [{\bf Q2}] How does $\mathtt{LRBM}$-$\mathtt{Boost}$ compare to other relational functional gradient boosting models? 
    \item [{\bf Q3}] Is an ensemble of weak relational regression trees more effective than a single strong relational regression tree for constructing Lifted RBMs?  
    \item [{\bf Q4}] Can we generate an interpretable lifted RBM from the ensemble of weak relational regression trees learned by {\tt LRBM-Boost}? 
\end{itemize}

\subsection{Experimental setup}
To answer these questions, we employ seven standard SRL data sets: 

\textbf{UW-CSE} \cite{mln} contains information about five university domains and the goal is to predict whether a student is $\fol{AdvisedBy}$ a professor. 

\textbf {IMDB} \cite{bottomupmln07} is a data set from movies domain that contains information about actors, directors and movies. The goal is to predict whether an actor $\fol{WorkedUnder}$ a director. 

\textbf{CORA} \cite{poon07} is a standard data set for citation matching that contains eight predicates about details of papers, their venues, and the authors. The aim is to predict whether two venues represent the $\fol{SameVenue}$. 

\textbf{SPORTS} is a data set garnered by crawling facts from NELL \cite{NELL2010} containing details about sports teams and their players. We aim to predict whether a team plays a particular sport (i.e. $\fol{TeamPlaysSport}$) in this domain. 

\textbf{MUTAGENESIS} \cite{lodhi2005} is data set that consists of information about molecules, their constituent atoms and their properties. The aim with this data set is to predict whether an atom is constituent in a molecule (i.e. $\fol{MoleculeContainsAtom}$). 

\textbf{YEAST2} \cite{pcrw10} contains facts about papers published between 1950 and 2012 about the yeast organism {\em Saccharomyces cerevisiae}. The target is whether a paper $\fol{Cites}$ another paper. Since this data is temporal, a recursive rule could potentially use the data from the future to predict the past. This requires restricting the data provided to the learning system from exposing any future data when predicting at the current time-step. 

\textbf{WEBKB} \cite{bottomupmln07} contains information about webpages of students, professors, courses etc. from four universities. We aim to predict whether a person is $\fol{CourseTA}$ of a given course.

For all data sets, we generate positive and negative examples in $1:2$ ratio, perform $5$-fold cross validation for every method being compared, and report AUC-ROC and AUC-PR on the resulting folds respectively. 

For all baseline methods, we use default settings provided by their respective authors. For our model, we learn $20$ RRTs, each with a maximum number of 4 leaf nodes. The learning rate of online coordinate descent was $0.05$. 

\subsection{Comparison of {\tt LRBM-Boost} to other relational neural models}
To answer {\bf Q1}, we compare our model to two recent relational neural models. The first baseline is Relational RBM ($\mathtt{RRBM}$-$\mathtt{C}$)  \cite{KaurEtAl18-RRBM}; this approach uses relational random walks to generate relational features that describe the structure of the domain. In fact, it propositionalizes and aggregates counts on these relational random walks as features to describe each training example. It should be noted that a key limitation of {\tt RRBM-C} is that it can only handle binary predicates; our approach {\tt LRBM-Boost} on the other hand,  can handle any arity.

The second baseline is Lifted Relational Neural Networks ($\mathtt{LRNN})$  \cite{LRNN2015}. $\mathtt{LRNN}$ mainly focuses on parameter optimization; the structure of the network is identified using a clause learner: $\mathtt{PROGOL}$ \cite{Muggleton1996}. $\mathtt{PROGOL}$ generated four, eight, six, three, ten, five rules for {\sc Cora}, {\sc Imdb}, {\sc Mutagenesis}, {\sc Sports}, {\sc Uw-Cse} and {\sc WebKB} respectively. As {\tt LRNN} cannot handle the temporal restrictions of {\sc Yeast2}, we do not evaluate {\tt LRNN} on it.

\begin{figure}[t]
    \centering
    \includegraphics[width=0.9\textwidth]{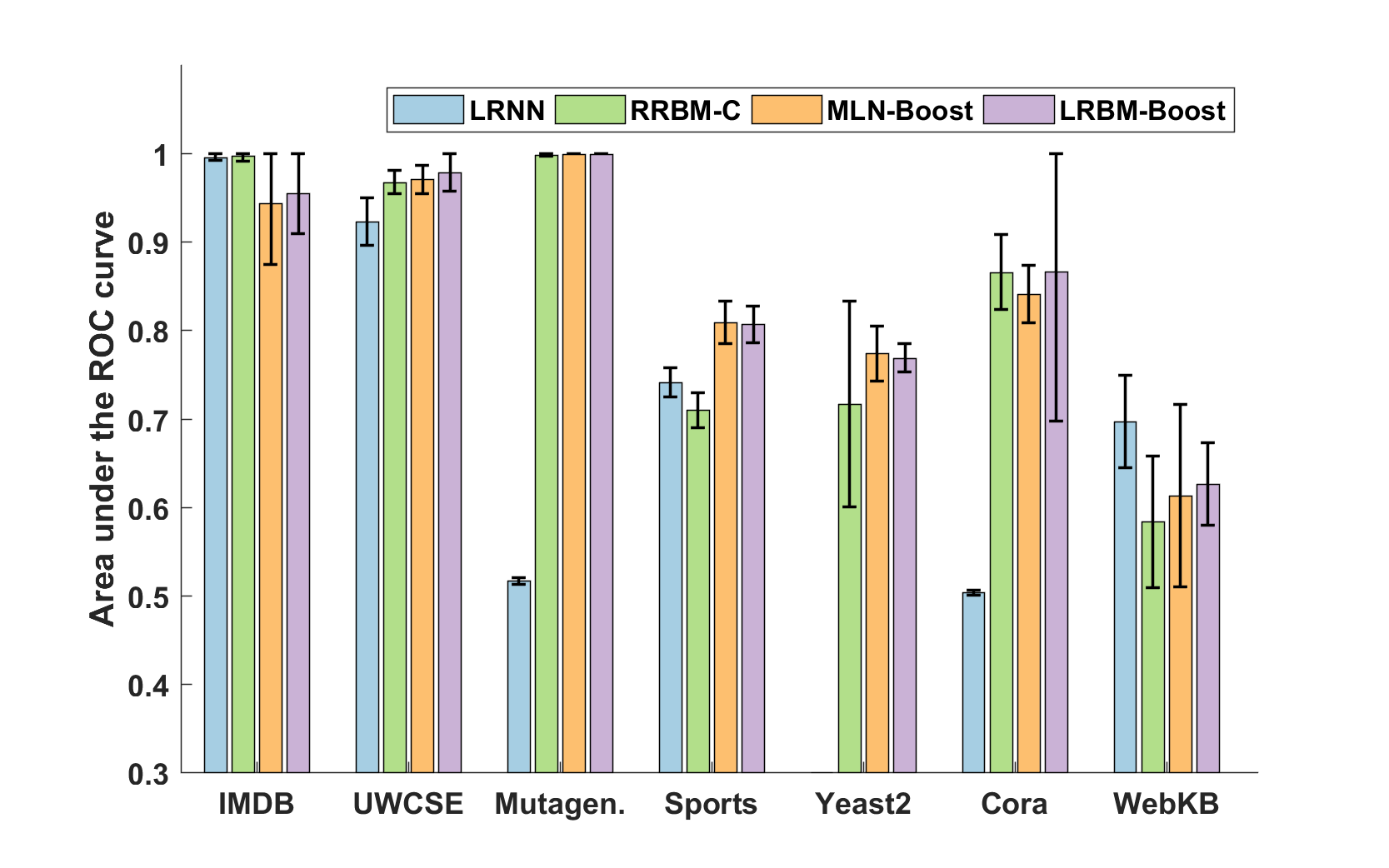}
    \caption{Comparing
    $\mathtt{LRNN}$,
    $\mathtt{RRBM}$-$\mathtt{C}$,
    $\mathtt{MLN}$-$\mathtt{Boost}$ and
    $\mathtt{LRBM}$-$\mathtt{Boost}$ on AUC-ROC.}
    \label{fig: Q1-AUC-ROC}
\end{figure}

\begin{figure}[t]
    \centering
    \includegraphics[width=0.9\textwidth]{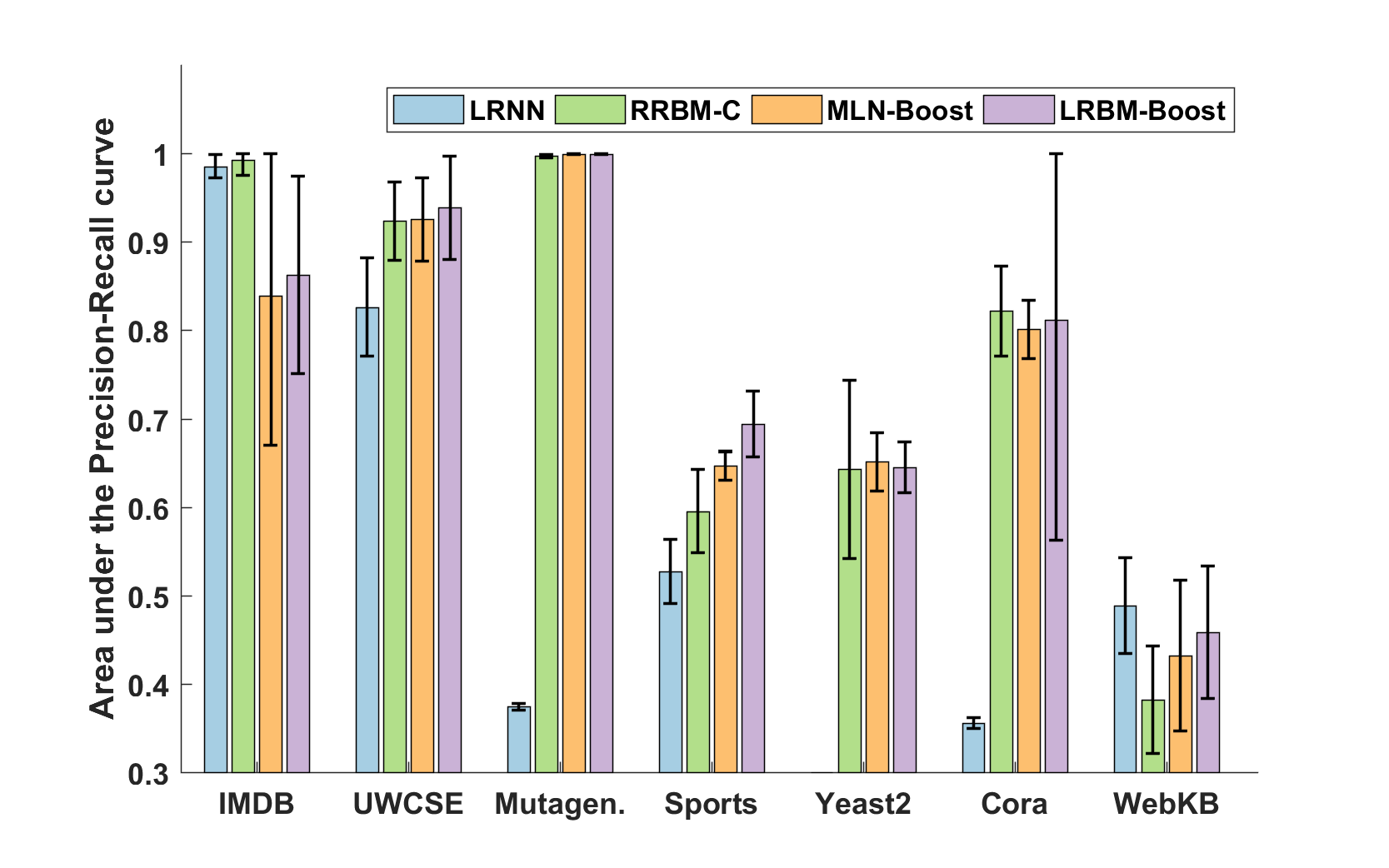}
    \caption{Comparing     
    $\mathtt{LRNN}$,
    $\mathtt{RRBM}$-$\mathtt{C}$,
    $\mathtt{MLN}$-$\mathtt{Boost}$ and
    $\mathtt{LRBM}$-$\mathtt{Boost}$ on AUC-PR.}
    \label{fig: Q1-AUC-PR}
\end{figure}

Figures \ref{fig: Q1-AUC-ROC} and \ref{fig: Q1-AUC-PR} present the results of this comparison on AUC-ROC and AUC-PR. $\mathtt{LRBM}$-$\mathtt{Boost}$ is significantly better than $\mathtt{LRNN}$ for {\sc Mutagenesis} and {\sc Cora} on both AUC-ROC and AUC-PR. Further, it also achieves better AUC-ROC and AUC-PR than $\mathtt{LRNN}$ on {\sc Sports} and {\sc Uw-Cse} data set. Compared to $\mathtt{RRBM}$-$\mathtt{C}$, $\mathtt{LRBM}$-$\mathtt{Boost}$ performs better for {\sc Sports} and {\sc WebKB} on both AUC-ROC and AUC-PR. Also, our proposed model performs better on {\sc Yeast2} on AUC-ROC. $\textbf{Q1}$ can be now be answered affirmatively: $\mathtt{LRBM}$-$\mathtt{Boost}$ either performs comparably to or outperforms state-of-the-art relational neural networks. 

\subsection{Comparison of {\tt LRBM-Boost} to other relational gradient-boosting models}
Since {\tt LRBM-Boost} is a relational neural network as well as a relational boosting model, we next compare it to two state-of-the-art relational functional gradient-boosting baselines: $\mathtt{MLN}$-$\mathtt{Boost}$ \cite{icdm11} and {\tt RDN-Boost} \cite{staraiBook}. Figures \ref{fig: Q1-AUC-ROC} and \ref{fig: Q1-AUC-PR} compare {\tt LRBM-Boost} to {\tt MLN-Boost}. $\mathtt{LRBM}$-$\mathtt{Boost}$ performs better than $\mathtt{MLN}$-$\mathtt{Boost}$ for {\sc Cora} and {\sc WebKB} on AUC-ROC metric. Further, it performs better than $\mathtt{MLN}$-$\mathtt{Boost}$ for {\sc Imdb}, {\sc Uw-Cse}, {\sc Sports} and {\sc WebKB} on AUC-PR. For all the other data sets, both the models have comparable performance. 

We compare {\tt LRBM-Boost} to {\tt RDN-Boost} in a separate experiment, owing to a key difference in experimental setting. We do not convert the arity of predicates to binary; rather, we compare {\tt RDN-Boost} and {\tt LRBM-Boost} by maintaining the original arity of all the predicates. The results of this experiment on four domains are reported in Table \ref{tab: RDN vs. RBM}. {\tt LRBM-Boost} outperforms {\tt RDN-Boost} in across the board, and substantially so on larger domains such as Cora. These comparisons allow us to answer {\bf Q2} affirmatively: {\tt LRBM-Boost} performs comparably or outperforms state-of-the-art SRL boosting baselines.

\begin{table*}[!t]
 \caption{Comparison of $\mathtt{LRBM}$-$\mathtt{Boost}$  and $\mathtt{RDN}$-$\mathtt{Boost}$.}
 \label{tab: RDN vs. RBM}
 \centering
\small
\begin{center}
\begin{tabular}{c|p{2.5cm}|p{2cm}|p{2cm}|p{2cm}} 
\Xhline{3\arrayrulewidth}
\Xhline{3\arrayrulewidth}
Data Set & Target & Measure & $\mathtt{LRBM}$-$\mathtt{Boost}$ & $\mathtt{RDN}$-$\mathtt{Boost}$ \\
\Xhline{3\arrayrulewidth}
\Xhline{3\arrayrulewidth}
\multirow{2}{4em}{\sc Uw$-$Cse} & \multirow{2}{4em}{\sc advisedBy} & AUC-ROC & 
0.9719 &         
0.9731 \\ [2pt]  
\cline{3-5}
& & AUC-PR & 
0.9158 &           
0.9049 \\ [2pt]    
\cline{3-5}
\Xhline{3\arrayrulewidth}
\multirow{2}{4em}{\sc Imdb} & \multirow{2}{4em}{\sc workedUnder} & AUC-ROC & 
0.9610 &           
0.9499 \\ [2pt]    
\cline{3-5}
& & AUC-PR & 
0.8789 &          
0.8537 \\ [2pt]   
\cline{3-5}
\Xhline{3\arrayrulewidth}
\multirow{2}{4em}{\sc Cora} & \multirow{2}{4em}{\sc SameVenue} & AUC-ROC & 
0.9469 &          
0.8985 \\ [2pt]   
\cline{3-5}
& & AUC-PR & 
0.9207 &           
0.8451 \\ [2pt]    
\cline{3-5}
\Xhline{3\arrayrulewidth}
\multirow{2}{4em}{\sc WebKB}& \multirow{2}{4em}{\sc courseTA} & AUC-ROC & 
0.6142 &           
0.6057 \\ [2pt]    
\cline{3-5}
& & AUC-PR & 
0.4553 &           
0.4490 \\ [2pt]    
\cline{3-5}
\Xhline{3\arrayrulewidth}
\end{tabular}
\end{center}
 \end{table*}

\subsection{Effectiveness of boosting relational ensembles}
To understand the importance of boosting trees to construct an LRBM, we compared the performance of the ensemble of relational trees learned by {\tt LRBM-Boost} to a {\em single relational tree},  similar to trees produced by the TILDE tree learner~\cite{TildeBlockeel1998,rpt2003}. For the latter, we learn a large lifted tree (of depth $10$), construct an RBM with the hidden layer being every path from root to leaf of this tree and refer to it as {\tt LRBM-NoBoost}.

Table \ref{tab:tableQ3} compares the performance of an ensemble (first row) vs. a single large tree (last row). {\tt LRBM-Boost} is statistically superior on {\sc Sports}, {\sc Yeast2} and {\sc Cora} on both AUC-ROC and AUC-PR and is comparable on others. This asserts the efficacy of learning ensembles of relational trees by LRBM-Boost rather than learning a single tree, thus affirmatively answering {\bf Q3}.

\begin{table*}[t]
 \caption{Comparison of (a) an ensemble of trees learned by $\mathtt{LRBM}$-$\mathtt{Boost}$,  (b) an explainable Lifted RBM constructed from the ensemble of trees learned by {\tt LRBM-Boost} and (c) learning a single, large, relational probability tree ({\tt LRBM-NoBoost}). 
 }
 \label{tab:tableQ3}
 \centering
\small
\begin{center}
\scalebox{0.9}{
\begin{tabular}{p{1.6cm}|p{0.7cm}|p{1.30cm}|p{1.30cm}|p{1.35cm}|p{1.30cm}|p{1.30cm}|p{1.30cm}} 
\Xhline{3\arrayrulewidth}
\Xhline{3\arrayrulewidth}
Model & AUC & Sports & IMDB & UW-CSE & Yeast2 & Cora & WebKB\\
\Xhline{3\arrayrulewidth}
\Xhline{3\arrayrulewidth}
\multirow{2}{4em}{\small Ensemble LRBM} & ROC & 
0.78$\pm$0.03 &          
0.95$\pm$0.05 &          
0.98$\pm$0.02 &          
0.77$\pm$0.02 &          
0.86$\pm$0.14 &          
0.63$\pm$0.05\\[2pt]    
\cline{2-8}
& PR & 
0.64$\pm$0.03 &         
0.86$\pm$0.11 &         
0.94$\pm$0.06 &         
0.64$\pm$0.03 &          
0.82$\pm$0.21 &          
0.46$\pm$0.08\\[2pt]    

\Xhline{3\arrayrulewidth}
\multirow{2}{4em}{\small Explainable LRBM} & ROC & 
0.75$\pm$0.01 &          
0.95$\pm$0.05 &          
0.95$\pm$0.04 &          
0.65$\pm$0.05 &          
0.80$\pm$0.19 &          
0.61$\pm$0.13\\[2pt]    
\cline{2-8}
& PR & 
0.57$\pm$0.01 &          
0.85$\pm$0.14 &          
0.89$\pm$0.06  &         
0.53$\pm$0.06 &          
0.70$\pm$0.29&          
0.46$\pm$0.10\\[2pt]    

\Xhline{3\arrayrulewidth}
\multirow{2}{4em}{\small NoBoost LRBM} & ROC & 
0.75$\pm$0.03&         
0.95$\pm$0.05&         
0.98$\pm$0.02&         
0.64$\pm$0.12 &        
0.75$\pm$0.21 &        
0.66$\pm$0.09\\[2pt]   
\cline{2-8}
& PR & 
0.61$\pm$0.01&        
0.86$\pm$0.11 &       
0.94$\pm$0.05 &       
0.50$\pm$0.14 &       
0.61$\pm$0.30 &       
0.48$\pm$0.07\\[2pt]  

 \Xhline{3\arrayrulewidth}
\Xhline{3\arrayrulewidth}
\end{tabular}
}
\end{center}
 \end{table*}
 
 \subsection{Interpretability of {\tt LRBM-Boost}} 
While {\bf Q1}--{\bf Q3} can be answered quantitatively, $\textbf{Q4}$ requires a qualitative analysis. It should be noted that boosted relational models (here, boosted LRBMs) learn and represent the underlying relational model as a sum of relational trees. When performing inference, this ensemble of trees is  {\em not converted to a large SRL model} as it is far more efficient to aggregate the predictions of the individual relational trees in the ensemble. 

\begin{figure}[!t]
\includegraphics[width=0.99\textwidth]{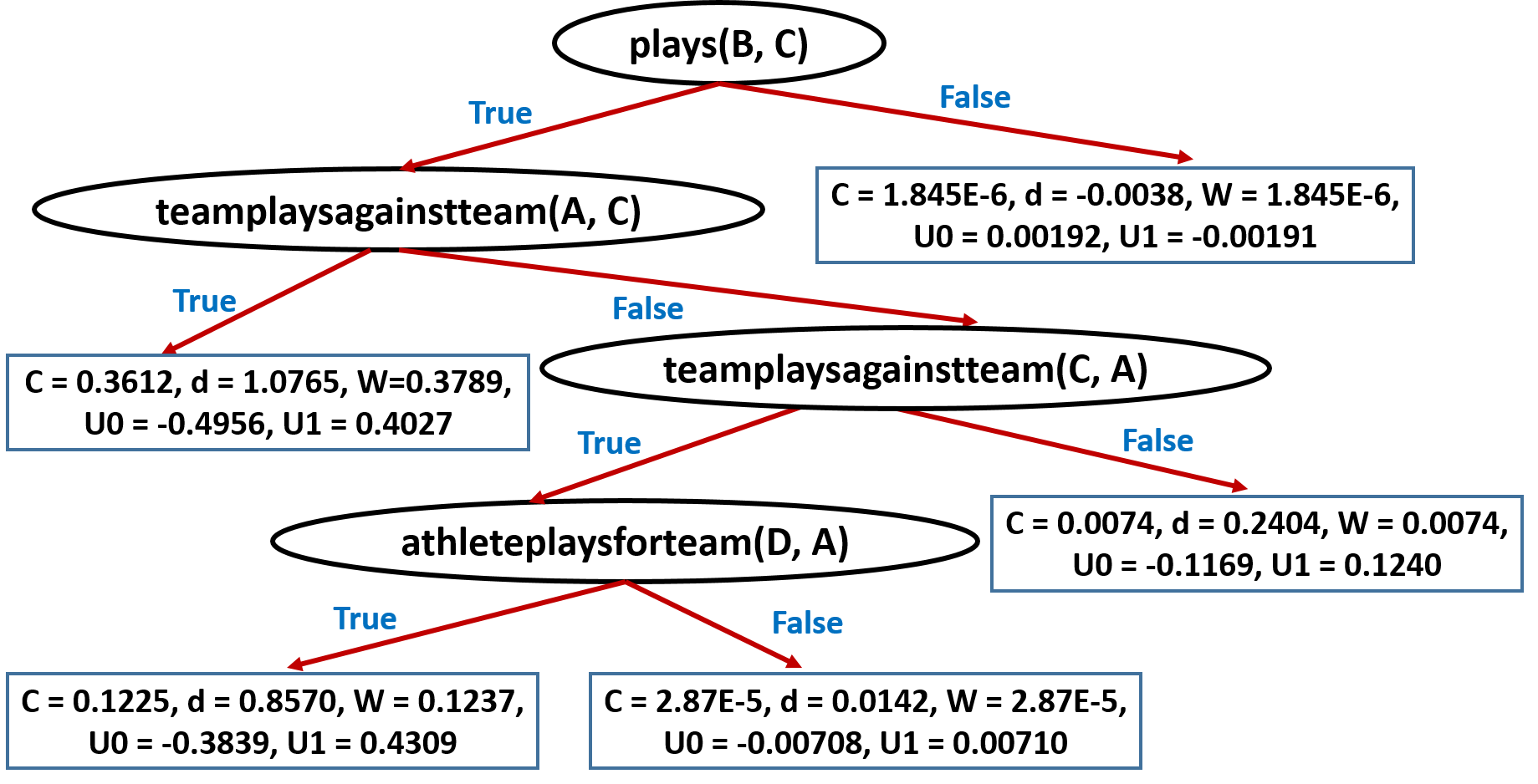}
\caption{{An example of a combined lifted tree learned from ensemble of trees. To construct this tree, we compute the regression value of each training example by traversing through all the boosted trees. Now a single large tree is overfit to this (modified) training set to generate a single tree.}}
\label{fig: CombinedTree}
\end{figure}

\begin{figure}[!t]
\includegraphics[width=0.99\textwidth]{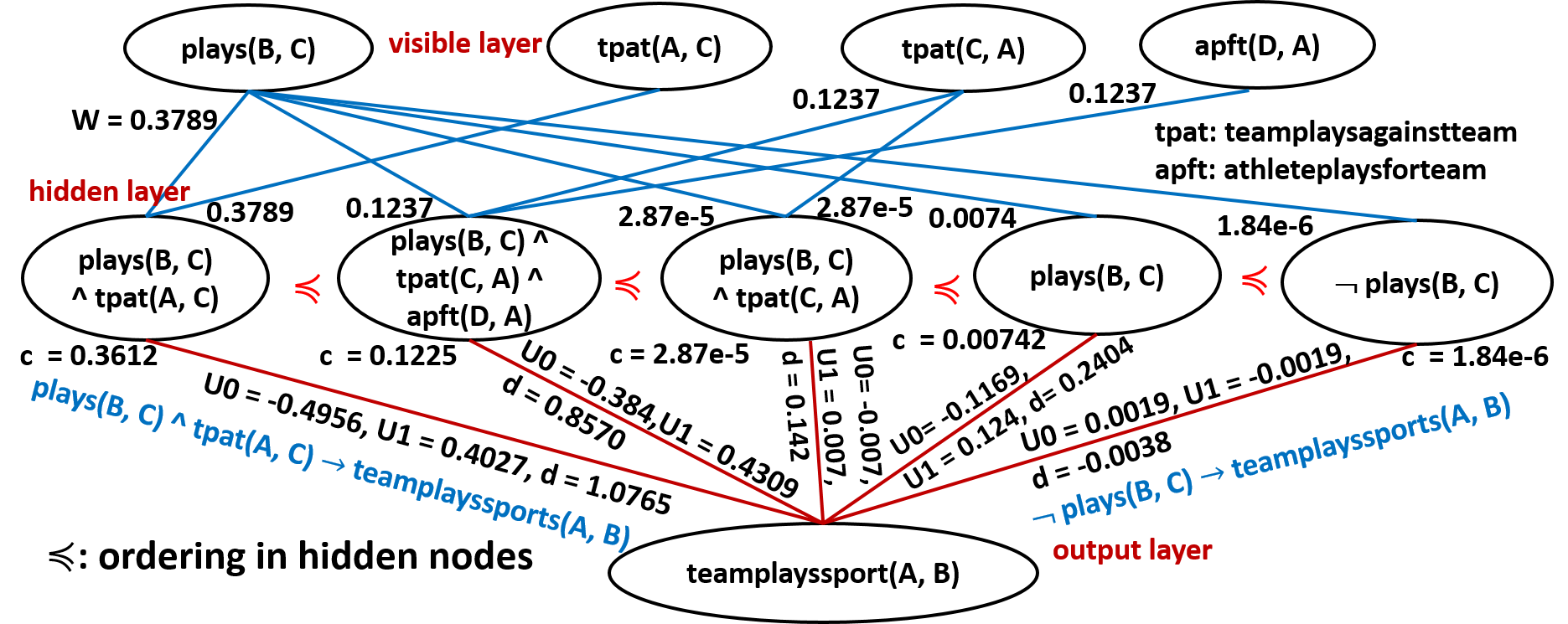}
\caption{Lifted RBM obtained from the combined tree in Figure~\ref{fig: CombinedTree}. Each path along the tree in that figure represents the corresponding hidden node of LRBM.}
\label{fig: RRBM}
\end{figure}

For {\tt LRBM-Boost}, however, it is possible to convert the ensemble-of-trees representation into a single LRBM. This step is typically performed to endow the model with interpretability, explainability or for relationship discovery. For {\tt LRBM-Boost}, this procedure is not exact, and the resulting single large LRBM is almost, but not exactly, equivalent to the ensemble of trees representation. The procedure itself is fairly straightforward: 
\begin{itemize}
\item learn a single RRT from the set of boosted RRTs~\cite{CravenAndShavlik1996} that make up the LRBM, that is, we empirically learn a single RRT by overfitting it to the predictions of the weak RRTs (Figure \ref{fig: CombinedTree}).
\item transform this single RRT to a lifted RBM (Figure \ref{fig: RRBM}); each path from root to leaf is a conjunction of relational features and enters the LRBM as a hidden node, with connections to all the output nodes and to the input nodes corresponding to the predicates that appear in that path.
\end{itemize}

This construction leads to sparsity as it allows for only one hidden node to be activated for each example. Of course, using clauses instead of trees as with boosting MLNs~\cite{icdm11}, could relax this sparsity as needed. For our current domains, this restriction does not significantly affect the performance as seen in Table~\ref{tab:tableQ3} showing the quantitative results of comparing the {\bf explainable} LRBM with the original {\bf ensemble} LRBM. There is no noticeable loss in performance as the AUC values decrease marginally, if at all. 

\begin{figure}
\includegraphics[width=0.9\textwidth]{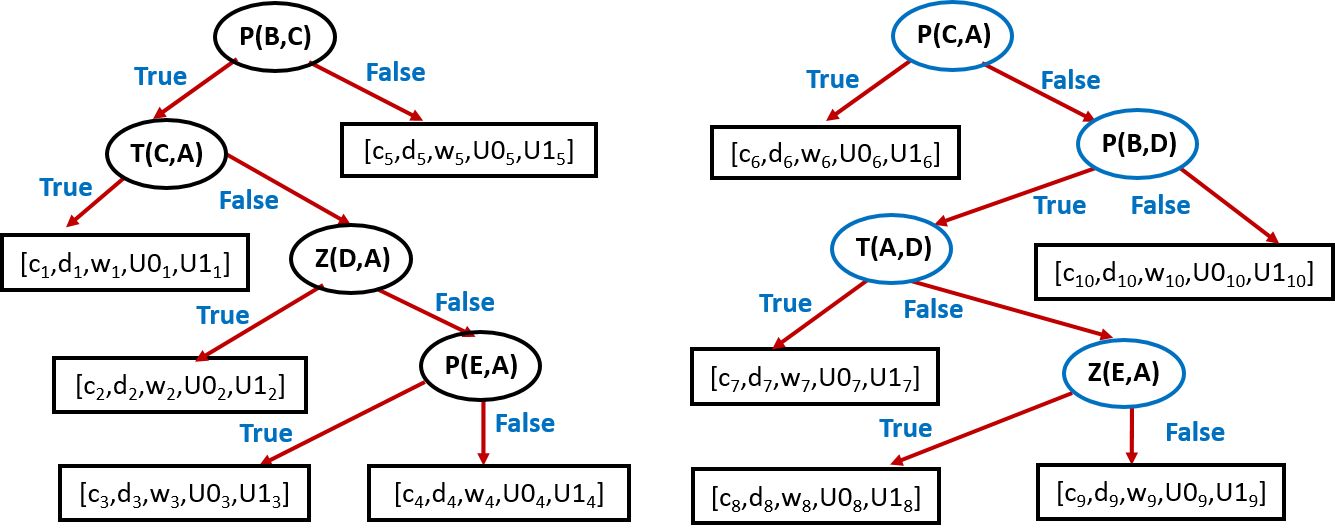}
\caption{{Ensemble of trees learned during training of LRBM-Boost. The ensemble of trees is generated in {\sc Sports} domain where predicate $\fol{P}$, $\fol{T}$, $\fol{Z}$ represent $\fol{plays(sports,team)}$, $\fol{teamplaysagainstteam(team,team)}$ and $\fol{athleteplaysforteam(athlete,team)}$ respectively and target $R$ represents $\fol{teamplayssport(team,sports)}$.}}
\label{fig: EnsembleOfTrees}
\end{figure}
\begin{figure}
\includegraphics[width=\textwidth]{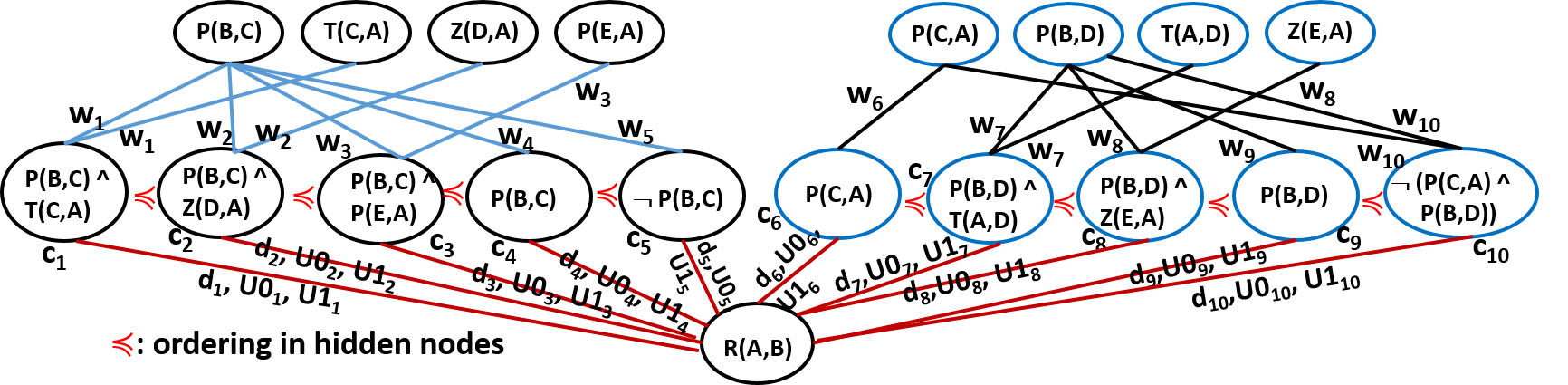}
\caption{Demonstration of the conversion of two lifted trees in Figure \ref{fig: EnsembleOfTrees} to LRBM. We create one hidden node for each path in each regression tree.}
\label{fig: EnsembleToRBMConversion}
\end{figure}



A simpler approach to constructing an explainable LRBM to skip aggregating the RRTs into a large tree and directly map {\em every path in every tree} to a hidden node in the LRBM. For instance, if the ensemble learned 20 balanced trees with $4$ paths in each of them, the resulting LRBM has $80$ lifted features. An example transformation is shown in Figure \ref{fig: EnsembleToRBMConversion} from two trees in Figure \ref{fig: EnsembleOfTrees}. Note that corresponding LRBM has $8$ hidden features which are conjunctions of the original trees. While in principle it results in an interpretable LRBM, this can result in a large number of hidden units and thus pruning strategies need to be employed, a direction that we will explore in the near future. In summary, it can be said that our LRBM is effective and explainable allowing when compared to the state-of-the-art approaches in several tasks.


\section{Conclusion}
We presented the first learning algorithm for learning the structure of a lifted RBM from data. Motivated by the success of gradient-boosting, our method learns a set of RRTs using boosting and then transforms them to a lifted RBM. The advantage of this approach is that it leads to learning a fully lifted model that is not propositionalized using any standard approaches. We also demonstrated how to induce a single explainable RBM from the ensemble of trees. Experiments on several data sets demonstrated the efficacy and effectiveness along with the explainability of the proposed approach. Combining the different trees in an analytical fashion is an interesting future direction. Enhancing the ability of the model to handle incomplete information is essential to adapt to real problems. Learning other distributions to learn truly hybrid models can lead to several adaptations on real data. Scaling to very large data sets (in the lines of relational embeddings) remains an exciting research direction. 

\section{Acknowledgement}
The authors gratefully acknowledge the support of AFOSR award FA9550-18-1-0462. Any opinions, findings, and conclusion or recommendations expressed in this material are those of the authors and do not necessarily reflect the view of the AFOSR or the United States government.

\section*{References}
\bibliography{RRBMBoost}

\begin{thebibliography}{10}
\expandafter\ifx\csname url\endcsname\relax
  \def\url#1{\texttt{#1}}\fi
\expandafter\ifx\csname urlprefix\endcsname\relax\def\urlprefix{URL }\fi
\expandafter\ifx\csname href\endcsname\relax
  \def\href#1#2{#2} \def\path#1{#1}\fi

\bibitem{RBMFirstPaper1987}
D.~E. {Rumelhart}, J.~L. {McClelland}, Information Processing in Dynamical
  Systems: Foundations of Harmony Theory, MIT Press, 1987, Ch. Parallel
  Distributed Processing: Explorations in the Microstructure of Cognition:
  Foundations.

\bibitem{ContrastiveDivergence2002}
G.~E. Hinton, Training products of experts by minimizing contrastive
  divergence, Neural Computation 14~(8).

\bibitem{PersistentCD2008}
T.~Tieleman, Training restricted boltzmann machines using approximations to the
  likelihood gradient, in: ICML, 2008.

\bibitem{paralleltempering2010}
G.~Desjardins, A.~Courville, Y.~Bengio, P.~Vincent, O.~Delalleau, Parallel
  tempering for training of restricted boltzmann machines, in: AISTATS, 2010.

\bibitem{taylor06}
G.~W. Taylor, G.~E. Hinton, S.~T. Roweis, Modeling human motion using binary
  latent variables, in: NeurIPS, 2007.

\bibitem{GraphConvolutionalNetwork2018}
M.~Schlichtkrull, T.~N. Kipf, P.~Bloem, R.~van~den Berg, I.~Titov, M.~Welling,
  Modeling relational data with graph convolutional networks, in: ESWC, 2018.

\bibitem{staraiBook}
L.~{De Raedt}, K.~Kersting, S.~Natarajan, D.~Poole, {Statistical Relational
  Artificial Intelligence: Logic, Probability, and Computation}, Morgan \&
  Claypool, 2016.

\bibitem{srlBook}
L.~Getoor, B.~Taskar, Introduction to Statistical Relational Learning, MIT
  Press, 2007.

\bibitem{CollectiveClassificationPham2017}
T.~Pham, T.~Tran, D.~Phung, S.~Venkatesh, Column networks for collective
  classification, in: AAAI, 2017.

\bibitem{RelNNKazemi2018}
S.~M. Kazemi, D.~Poole, Relnn: {A} deep neural model for relational learning,
  in: AAAI, 2018.

\bibitem{LRNN2015}
G.~\v{S}ourek, V.~Aschenbrenner, F.~\v{Z}elezny, O.~Ku\v{z}elka, Lifted
  relational neural networks, in: COCO, 2015.

\bibitem{KaurEtAl18-RRBM}
N.~Kaur, G.~Kunapuli, T.~Khot, K.~Kersting, W.~Cohen, S.~Natarajan, Relational
  restricted boltzmann machines: A probabilistic logic learning approach, in:
  ILP, 2017.

\bibitem{Friedman2001}
J.~Friedman, Greedy function approximation: A gradient boosting machine, Annals
  of Statistics.

\bibitem{ImitationLearning2011}
S.~Natarajan, S.~Joshi, P.~Tadepalli, K.~Kersting, J.~Shavlik, Imitation
  learning in relational domains: A functional-gradient boosting approach, in:
  IJCAI, 2011.

\bibitem{icdm11}
T.~Khot, S.~Natarajan, K.~Kersting, J.~Shavlik, Learning {M}arkov logic
  networks via functional gradient boosting, in: ICDM, 2011.

\bibitem{rdnmlj11}
S.~Natarajan, T.~Khot, K.~Kersting, B.~Guttmann, J.~Shavlik, Gradient-based
  boosting for statistical relational learning: The relational dependency
  network case, MLJ.

\bibitem{TildeCRF2006}
B.~Gutmann, K.~Kersting, Tildecrf: Conditional random fields for logical
  sequences, in: ECML, 2006.

\bibitem{DiscrRBM2008}
H.~Larochelle, M.~Mandel, R.~Pascanu, Y.~Bengio, Learning algorithms for
  classification restricted boltzmann machines, JMLR.

\bibitem{TildeBlockeel1998}
H.~Blockeel, L.~De~Raedt, Top-down induction of first-order logical decision
  trees, Artificial Intelligence 101.

\bibitem{RLRBoost2018}
N.~Ramanan, G.~Kunapuli, T.~Khot, B.~Fatemi, S.~M. Kazemi, D.~Poole,
  K.~Kersting, S.~Natarajan, Structure learning for relational logistic
  regression: An ensemble approach, in: KR, 2018.

\bibitem{Yang2016AAAI}
S.~Yang, T.~Khot, K.~Kersting, S.~Natarajan, Learning continuous-time
  {B}ayesian networks in relational domains: A non-parametric approach, in:
  AAAI, 2016.

\bibitem{RESCAL2011}
M.~Nickel, V.~Tresp, H.-P. Kriegel, A three-way model for collective learning
  on multi-relational data, in: ICML, 2011.

\bibitem{TransE2013}
A.~Bordes, N.~Usunier, A.~Garcia-Duran, J.~Weston, O.~Yakhnenko, Translating
  embeddings for modeling multi-relational data, in: NeurIPS, 2013.

\bibitem{NTN2013}
R.~Socher, D.~Chen, C.~D. Manning, A.~Ng, Reasoning with neural tensor networks
  for knowledge base completion, in: NeuRIPS, 2013.

\bibitem{DistMult2015}
B.~Yang, W.~Yih, X.~He, J.~Gao, L.~Deng, Embedding entities and relations for
  learning and inference in knowledge bases, in: ICLR, 2015.

\bibitem{HolE2016}
M.~Nickel, L.~Rosasco, T.~Poggio, Holographic embeddings of knowledge graphs,
  in: AAAI, 2016.

\bibitem{complex2016}
T.~Trouillon, J.~Welbl, S.~Riedel, E.~Gaussier, G.~Bouchard, Complex embeddings
  for simple link prediction, in: ICML, 2016.

\bibitem{CLIPPLusPlus2014}
M.~V. Fran\c{c}a, G.~Zaverucha, A.~S. D'avila~Garcez, Fast relational learning
  using bottom clause propositionalization with artificial neural networks, MLJ
  94~(1).

\bibitem{DiMaio2004}
F.~DiMaio, J.~Shavlik, Learning approximation to inductive logic programming
  clause evaluation, in: ILP, 2004.

\bibitem{DRM2013}
H.~Lodhi, Deep relational machines, in: ICONIP, 2013.

\bibitem{DeepMindNN2017}
A.~Santoro, et~al., A simple neural network module for relational reasoning,
  in: NeurIPS, 2017.

\bibitem{FewShotLearning2018}
F.~Sung, Y.~Yang, L.~Zhang, T.~Xiang, P.~H.~S. Torr, T.~M. Hospedales, Learning
  to compare: Relation network for few-shot learning, in: CVPR, 2018.

\bibitem{RelNNforObjectDetection2018}
H.~Hu, J.~Gu, Z.~Zhang, J.~Dai, Y.~Wei, Relation networks for object detection,
  in: CVPR, 2018.

\bibitem{NiepertGraphConNet2016}
M.~Niepert, M.~Ahmed, K.~Kutzkov, Learning convolutional neural networks for
  graphs, in: ICML, 2016.

\bibitem{GraphNeuralNetwork2009}
F.~Scarselli, M.~Gori, A.~C. Tsoi, M.~Hagenbuchner, G.~Monfardini, The graph
  neural network model, IEEE Transactions.

\bibitem{GatedCHBM2015}
Y.~Huang, W.~Wang, L.~Wang, Conditional high-order boltzmann machine: A
  supervised learning model for relation learning, in: ICCV, 2015.

\bibitem{LRBM2014}
K.~Li, J.~Gao, S.~Guo, N.~Du, X.~Li, A.~Zhang, {LRBM}: A restricted boltzmann
  machine based approach for representation learning on linked data, in: ICDM,
  2014.

\bibitem{Muggleton94}
S.~Muggleton, L.~D. Raedt, Inductive logic programming: Theory and methods,
  Journal Of Logic Programming 19.

\bibitem{Wright15-CoordinateDescent}
S.~J. Wright, Coordinate descent algorithms, Mathematical Programming 151~(1)
  (2015) 3--34.

\bibitem{mln}
M.~Richardson, P.~Domingos, {M}arkov logic networks, MLJ.

\bibitem{bottomupmln07}
L.~Mihalkova, R.~Mooney, Bottom-up learning of {Markov logic} network
  structure, in: ICML, 2007.

\bibitem{poon07}
H.~Poon, P.~Domingos, Joint inference in information extraction, in: AAAI,
  2007.

\bibitem{NELL2010}
A.~Carlson, J.~Betteridge, B.~Kisiel, B.~Settles, E.~R. Hruschka, Jr., T.~M.
  Mitchell, Toward an architecture for never-ending language learning, in:
  AAAI, 2010.

\bibitem{lodhi2005}
H.~Lodhi, S.~Muggleton, Is mutagenesis still challenging ?, in: ILP, 2005.

\bibitem{pcrw10}
N.~Lao, W.~Cohen, Relational retrieval using a combination of path-constrained
  random walks, JMLR 81.

\bibitem{Muggleton1996}
S.~Muggleton, Learning from positive data, in: ILP, 1997.

\bibitem{rpt2003}
J.~Neville, D.~Jensen, L.~Friedland, M.~Hay, Learning relational probability
  trees, in: KDD, 2003.

\bibitem{CravenAndShavlik1996}
M.~W. Craven, J.~W. Shavlik, Extracting tree-structured representations of
  trained networks, in: NeurIPS, 1995.

\end{thebibliography}

\section*{Appendix: Inference in a Lifted RBM}



The lifted RBM is a template that is grounded for each example during inference. We first unify the example with the head of the clause (present at the output layer of LRBM), to obtain a partial grounding of the body of the clause. The full grounding is then obtained by unifying the partially-ground clause with evidence to find at least one instantiation of the body of the clause.  We illustrate the inference procedure for a Lifted RBM with three hidden nodes, and each hidden node corresponding to the rules ($h_1$)--($h_3$).  

\begin{figure}[!thb]
    \centering
    \includegraphics[scale=0.7]{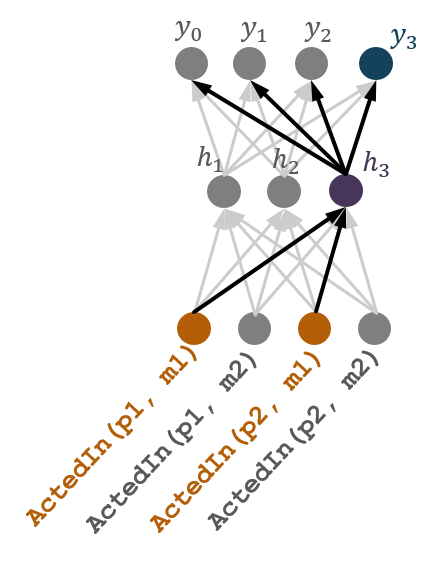}
    \caption{LRBM inference for Example 1.}
    \label{fig: lrbm inference ex1}
\end{figure}

\paragraph{Example 1} We are given  facts: $\fol{ActedIn(p1, m1)}$, $\fol{ActedIn(p1, m2)}$, $\fol{ActedIn(p2,}$ $\fol{m1)}$, $\fol{ActetdIn(p2, m2) }$. The number of substitutions depends on the query. Let us assume that the query is $\fol{Collab(p1, p2)}$ (did $\fol{p1}$ and $\fol{p2}$ collaborate?), which results in the partial substitution: $\theta = \{ \fol{P_1/p1, P_2/p2} \}$. The inference procedure will proceed as follows:

\begin{itemize}
\item The bodies of the clauses ($h_1$)--($h_3$) are partially grounded using $\theta = \{ \fol{P_1/p1, P_2/p2} \}$:
\[
\begin{array}{l@{\hspace{0.35in}}c}
\left( \begin{array}{r}
    \fol{DirectedBy(M_1, p1) \wedge InGenre(M_1, G_1) \, \wedge } \\
    \fol{ActedIn(p2, M_2) \wedge InGenre(M_2, G_2) \, \wedge }\\ 
    \neg \, \fol{SameGenre(G_1, G_2)} \,\,\,\,\,\,\\
\end{array}\right)  & (h_1) \\
\fol{DirBy(M_1, p1) \wedge ActedIn(P_3, M_1) \wedge SamePerson(P_3, p2)} & (h_2)\\
\fol{ActedIn(p1, M) \wedge ActedIn(p2, M)}. & (h_3)
\end{array}
\]

\item Next, since the facts do not contain any information about $\fol{DirectedBy}$ or $\fol{SamePerson}$, $h_1$ and $h_2$ will not be satisfied. 

\item In order to prove the satisfiability of $h_3$, we look at all the available facts as we attempt to unify each fact with the partially-grounded clause. Say we first unify $\fol{ActedIn}$($\fol{p1}$, $\fol{m1}$) with $h_3$ that gives us:
\[
\fol{ActedIn}(\fol{p1}, \fol{m1}), \fol{ActedIn}(\fol{p2}, \fol{M}),  \hspace{0.25in} (h_3)
\]
resulting in the grounding: $\theta = \{\fol{P_1/p1, M/m1, P_2/p2}\}$. The second fact $\fol{ActedIn}$($\fol{p1}$, $\fol{m2}$) does not unify with this partially-grounded clause. However, the third fact $\fol{ActedIn}(\fol{p2}$, $\fol{m1})$ unifies with $h_3$ giving us a fully-grounded clause: 
\[
\fol{ActedIn}(\fol{p1}, \fol{m1}), \fol{ActedIn}(\fol{p2}, \fol{m1}).  \hspace{0.25in} (h_3)
\]
The input nodes corresponding to the unified facts $\fol{ActedIn(p1, m1),  Acted}$ $\fol{In(p2, m1) }$ are activated. As soon as this clause is satisfied the search terminates. The main conclusion to be drawn is that as soon as the clause is satisfied once, model does not check for another satisfaction and terminates the search by returning true.

\item The inputs are then propagated through the RBM, and the class output probabilities are computed based on the RBM edge parameters/weights. The activation paths for inference given this query and facts are shown in Figure \ref{fig: lrbm inference ex1}.
\end{itemize}

\paragraph{Example 2} We are given facts: $\fol{DirectedBy(m1, p1)}$, $\fol{InGenre(m1, g1)}$, $\fol{ActedIn}$ $\fol{(p2, m2)}$,   $\fol{InGenre(m2, g2)}$, $\fol{DirectedBy(m01, p01)}$, $\fol{ActedIn(p03, m01)}$,\\ \noindent $\fol{SamePerson(p03, p02)}$. Recall that the number of substitutions depends on the query. Let us assume that the query is $\fol{Collab(p01, p02)}$ (did $\fol{p01}$ and $\fol{p02}$ collaborate?), which results in the partial substitution: $\theta = \{ \fol{P_1/p01, P_2/p02} \}$. The inference procedure will proceed as follows:

\begin{itemize}
\item The bodies of the clauses ($h_1$)--($h_3$) are partially grounded using $\theta = \{ \fol{P_1/p01, P_2/p02} \}$:
\[
\begin{array}{l@{\hspace{0.35in}}c}
\left( \begin{array}{r}
    \fol{DirectedBy(M_1, p01) \wedge InGenre(M_1, G_1) \, \wedge } \\
    \fol{ActedIn(p02, M_2) \wedge InGenre(M_2, G_2) \, \wedge }\\ 
    \neg \, \fol{SameGenre(G_1, G_2)} \,\,\,\,\,\,\\
\end{array}\right)  & (h_1) \\
\fol{DirBy(M_1, p01) \wedge ActedIn(P_3, M_1) \wedge SamePerson(P_3, p02)}  &
(h_2)\\
\fol{ActedIn(p01, M) \wedge ActedIn(p02, M)}. &
(h_3)
\end{array}
\]

\item Unifying the partially-grounded clauses with the facts, we will have that $h_1$ and $h_3$ will not be satisfied. However, unification yields one fully-grounded $h_2$ will:
\[
\fol{DirBy(m01, p01) \wedge ActedIn(p03, m01) \wedge SamePerson(p03, p02)},  \hspace{0.3in} (h_2)
\]
which has the substitution: $\theta = \{\fol{P_1/p01, M/m01, P_2/p02, P_3/p03}\}$. As before, once a satisfied grounding is obtained, the search is terminated.

\item The RBM is unrolled as in Example 1, and the appropriate facts that appear in this grounding are activated in the input layer. The prediction is obtained by propagating these inputs through the network.
\end{itemize}

\end{document}